\documentclass[letterpaper]{article} 
\usepackage{aaai2026}  
\nocopyright
\usepackage{times}  
\usepackage{helvet}  
\usepackage{courier}  
\usepackage[hyphens]{url}  
\usepackage{graphicx} 
\usepackage{natbib}  
\usepackage{caption} 
\usepackage{algorithm}
\usepackage{algorithmic}

\usepackage{amsmath}

\usepackage{newfloat}
\usepackage{listings}
\DeclareCaptionStyle{ruled}{labelfont=normalfont,labelsep=colon,strut=off} 
\floatstyle{ruled}
\newfloat{listing}{tb}{lst}{}
\floatname{listing}{Listing}

  \usepackage{multirow}
  \usepackage{array}

  \usepackage{subfigure}
\usepackage{subcaption}
\usepackage{amsthm}
\usepackage{amssymb}
\usepackage{booktabs}
\usepackage{xcolor}

\newtheorem{theorem}{Theorem}

\newtheorem{definition}{Definition}

\title{A Generalized Learning Framework for Self-Supervised Contrastive Learning}

\author{
    Lingyu Si\textsuperscript{\rm 1,\rm 2},
    Jingyao Wang\textsuperscript{\rm 1,\rm 2},
    Wenwen Qiang\textsuperscript{\rm 1,\rm 2}\thanks{Corresponding author.},
}
\affiliations{

    \textsuperscript{\rm 1}University of Chinese Academy of Sciences,
    \textsuperscript{\rm 2}Institute of Software Chinese Academy of Sciences\\
}

\usepackage{bibentry}

\begin{document}

\maketitle

\begin{abstract}
Self-supervised contrastive learning (SSCL) has recently demonstrated superiority in multiple downstream tasks. In this paper, we generalize the standard SSCL methods to a \textbf{G}eneralized \textbf{L}earning \textbf{F}ramework (GLF) consisting of two parts: the aligning part and the constraining part. We analyze three existing SSCL methods: BYOL, Barlow Twins, and SwAV, and show that they can be unified under GLF with different choices of the constraining part. We further propose empirical and theoretical analyses providing two insights into designing the constraining part of GLF: intra-class compactness and inter-class separability, which measure how well the feature space preserves the class information of the inputs. However, since SSCL can not use labels, it is challenging to design a constraining part that satisfies these properties. To address this issue, we consider inducing intra-class compactness and inter-class separability by iteratively capturing the dynamic relationship between anchor and other samples and propose a plug-and-play method called \textbf{A}daptive \textbf{D}istribution \textbf{C}alibration (ADC) to ensure that samples that are near or far from the anchor point in the original input space are closer or further away from the anchor point in the feature space. Both the theoretical analysis and the empirical evaluation demonstrate the superiority of ADC.
\end{abstract}


\section{Introduction}
\label{sec:intro}
Representation learning without pairwise constraints is a high-profile research concern in the field of machine learning. Contrastive learning (CL), as an innovative self-supervised contrastive learning (SSCL) approach, has recently demonstrated superiority in various tasks, e.g., classification, object detection, and segmentation \citep{jaiswal2020survey,SI2022727,wang2023amsa,radford2021learning,wang2024image}. A characteristic of CL is its instance-based learning paradigm. That is, CL considers each sample in the training dataset as a separate class. Based on such a simple but graceful paradigm, CL can learn semantic information from the data itself, which can be delivered to various downstream tasks with satisfactory performance. As CL continues to develop, state-of-the-art methods such as SimCLR \citep{chen2020simple} and BYOL \citep{byol} are narrowing the performance gap with supervised methods.

In general, CL learns the feature extractor by minimizing the contrastive loss. \citep{wang2020understanding} decomposed the contrastive loss into two parts: alignment and uniformity. Alignment aims to constrain the similarity between positive samples and anchors, while uniformity aims to constrain the distribution of all samples to satisfy a uniform distribution. \citep{chen2021intriguing} provided a closer analysis of uniformity and suggested that it is also valid to constrain the distribution of all samples to satisfy other distributions, such as Gaussian or high-dimensional uniform distributions. Among these methods, the first term in their objective can be interpreted as aligning, while the second term is considered as constraining the data distribution to satisfy a definitive condition. From this perspective, we generalize the standard self-supervised contrastive learning (SSCL) method and propose a \textbf{G}eneralized \textbf{L}earning \textbf{F}ramework (GLF). Specifically, it consists of two parts: the aligning part and the constraining part. The aligning part aims to align two augmented samples with the same ancestor in the feature space, while the constraining part imposes a priori constraints on the training data. GLF is generalized in the sense that its constraining part can be any reasonable form of constraints, not limited to distributional constraints.

One question that arises is what kind of constraining part can be beneficial for improving the performance of contrastive learning in downstream tasks. We answer this question from two aspects. The first aspect is empirical analysis (\textbf{Subsection ``Empirical Analysis''}). In brief, we provide an empirical analysis of four representative SSCL methods and conclude that their objective functions do not model the dynamic relationship between augmented samples. As a result, they all face the problem of insufficient separability of data distributions of different categories. We verify this by running toy experiments on four benchmark datasets with five different constraints, which further indicate that if the constraints under the new framework can make points from different categories move away from each other, it can significantly improve the performance of SSCL methods. The second aspect is theoretical analysis (\textbf{Subsection ``Theoretical Analysis''}). We further demonstrate that if the constraining part under the new framework can make points of the same category in the feature space cluster with each other, it can significantly improve the performance of SSCL methods. These two aspects provide insights into the design of the constraining part under the new framework. An ideal constraining part should make points of the same class cluster together as much as possible while keeping points of different classes as far apart as possible.

Guided by the above insights, we propose a new constraining part under GLF called \textbf{A}daptive \textbf{D}istribution \textbf{C}alibration (ADC). ADC can be viewed as a learnable regularization term that can be directly integrated into existing SSCL models. The core idea of ADC is to iteratively treat the samples in the dataset as anchors and control other samples to aggregate or move away from anchors according to the similarity between samples. This induces the effect of aggregating samples with the same classes and separating samples with dissimilar classes. ADC consists of two parts: the Distribution Calibration Module (DCM) and the Local Preserving Module (LPM). DCM aims to aggregate similar samples together and separate dissimilar samples from each other through learnable distribution calibration. Specifically, DCM assumes that there are two distributions in the feature space: a calibration distribution which is set as a multivariate Gaussian distribution, and a data distribution which is set as a multivariate $t$-distribution. By minimizing the KL-divergence between these two distributions, samples in the feature space can be automatically aggregated or separated. However, the poor performance of the feature extractor at the early stage of training may cause the distance of the samples in the feature space from the anchor point to change, which in turn causes the DCM to aggregate or separate the samples incorrectly. Meanwhile, when outliers are selected as anchor, DCM will cause some samples in the dataset to cluster towards outliers, resulting in wrong clustering among samples.
To alleviate these problems, LPM proposes to constrain the positions of samples in the feature space to be consistent with their positions in the original space using a Dirichlet distribution (DD). Also, LPM uses an interesting observation of DD to mitigate the effects of outliers. Our major contributions are as follows:
\begin{itemize}
    \item We develop a generalized learning framework (GLF) that extends the standard SSL approach. GLF consists of two components: an aligning part and a constraining part. 
    \item Based on the empirical and theoretical analyses, we find that an ideal constraint should maximize the clustering of points belonging to the same class while keeping points from different classes as far apart as possible.
    \item We propose a plug-and-play method that can be integrated into existing models. The core concept is to ensure points that are close or far away in the original input space are closer or farther away in the feature space.
    \item We provide a theoretical guarantee on the error bound for the classification task and conduct empirical evaluations to demonstrate that our proposed method can improve the performance of various state-of-the-art SSCL methods.
\end{itemize}

\begin{figure}[!htb]
    \centering
    \subfigure[CIFAR-10]{\includegraphics[width=0.23\textwidth]{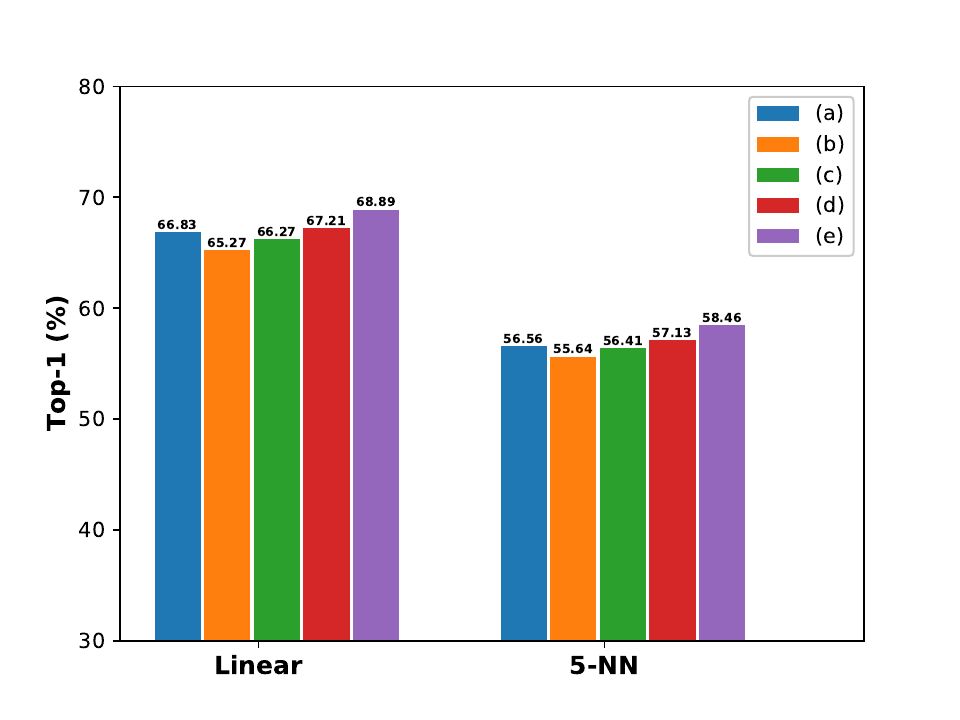}\label{fig: sub_figure1}}
    \subfigure[STL-10]{\includegraphics[width=0.23\textwidth]{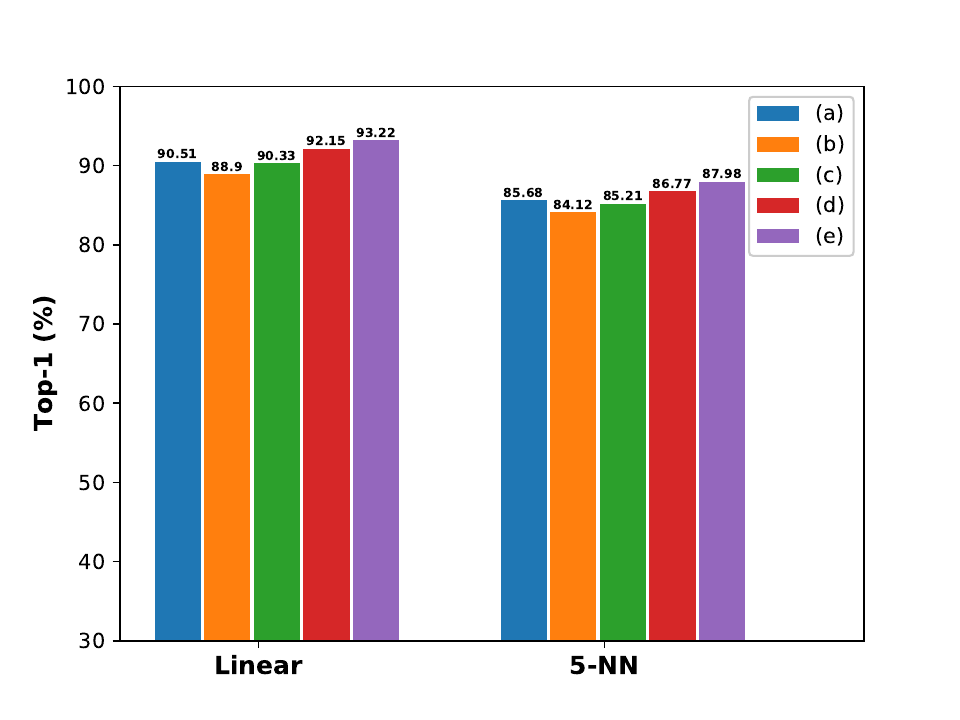}\label{fig: sub_figure2}}
    \caption{Motivating Example. We have experimented with five different approaches on two benchmark datasets. Different colors indicate the comparison learning models obtained by training based on different prior distributions.}
    \label{fig: fig1}
\end{figure}

\section{Related Work}
Contrastive learning (CL) has been well studied as a high-profile approach to learning visual representations without labels \citep{oord2018representation, tian2020contrastive, hjelm2018learning}. SimCLR \citep{chen2020simple} is the first widely used contrastive learning method that significantly narrows the gap with supervised learning. However, SimCLR requires large training batches and, thus, high computational resource requirements. Then, MoCo \citep{moco, chen2020improved, chen2021empirical} is proposed to mitigate this problem by dynamic memory allocation. MetAug \citep{li2022metaug} proposes an augmentation generation mechanism to generate hard positive samples to alleviate the problem of too many negative samples. Also, some methods propose to alleviate this problem by using a no-negative-sample strategy, e.g., BYOL \citep{grill2020bootstrap}, W-MSE \citep{ermolov2021whitening}, Simsiam \citep{chen2021exploring}, and Barlow Twins \citep{zbontar2021barlow}. A problem among these methods is that the intrinsic structure of the data distribution is ignored. To solve this problem, SwAV \citep{swav} and PCL \citep{li2020prototypical} propose to excavate the clustering structures embedded in data distribution. LMCL \citep{chen2021large} proposes to mine the large margin between the positive samples and the negative samples related to the anchor. Contrastive learning can be viewed as an instance-based learning paradigm. Thus, the related methods can not explore the relationship between different instances. Based on this problem, ReSSL \citep{zheng2021ressl, tomasev2022pushing} is proposed to measure the similarity of the data distribution based on two augmented samples. Different from these mentioned methods, the proposed ADC is to induce the clustering structures in an instance-based manner. We aim to constrain points closer to the anchor in the original space to be closer to the feature space and points farther from the anchor to be farther away in the feature space.

\section{Problem Formulation and Analysis}

In this section, we present an overview of SimCLR \citep{chen2020simple} for the alignment and uniformity properties and propose a generalized framework by considering the uniformity of SimCLR as a constraint on the training data. We then revisit three representative SSCL methods within this framework. Furthermore, we provide insights into SSCL through both theoretical and empirical analysis.

Given a mini-batch of training data ${X_{tr}} = \left\{ {{x_i}} \right\}_{i = 1}^N$, where ${{x_i}}$ denotes the $i$-th sample and $N$ represents the number of samples, we generate an augmented dataset $X_{tr}^{aug} = \left\{ {x_1^1,x_1^2,...,x_N^1,x_N^2} \right\}$ by applying stochastic data augmentation (e.g., random crop) to transform each sample $x_i$ into two augmented views $x^1_i$ and $x^2_i$. We can also obtain $X_{tr}^{aug} = \left\{ {{}^1X_{tr}^{aug},{}^2X_{tr}^{aug}} \right\}$, where ${}^iX_{tr}^{aug} = \left\{ {x_1^i,...,x_N^i} \right\}$ and $i \in \left\{ {1,2} \right\}$. The samples in ${X_{tr}}$ are considered as ancestors of those in $X_{{tr}}^{aug}$. SSCL \citep{chen2020simple, wang2020understanding} involves training a feature extractor $f$ to project input samples from their original space $\mathcal{X}$ into a feature space $\mathcal{Z}$, thus we have: $z_i^j = f( {x_i^j} )$, where $i \in \left\{ {1,2,...,N} \right\}$ and $j \in \left\{ {1,2} \right\}$. The goal of SSCL is to learn a general representation that is suitable for downstream tasks.

\subsection{Overview of SimCLR}

SimCLR \citep{chen2020simple} is an instance-based learning method that treats each sample as a separate class during training. It minimizes the contrastive loss in the feature space to bring points of the same class closer together while separating points of different classes. The objective is:
\begin{equation}\label{eq:NCE}
\begin{array}{l}
\scalebox{0.86}{${{\mathcal{L}}_{{\rm{NCE}}}} = 
\sum\limits_{i = 1,l = 1}^{i = N,l = 2} { - \log {\textstyle{{\exp \left[ {{\textstyle{{{\rm{sim}}\left( {z_i^l,z_i^{3 - l}} \right)} \over \tau }}} \right]} \over {\exp \left[ {{\textstyle{{{\rm{sim}}\left( {z_i^l,z_i^{3 - l}} \right)} \over \tau }}} \right] + \sum\limits_{j = 1,j \ne i,k = 1}^{j = N,k = 2} {\exp \left[ {{\textstyle{{{\rm{sim}}\left( {z_i^l,z_j^k} \right)} \over \tau }}} \right]} }}}} ,$}
\end{array}
\end{equation}
where $\tau$ represents the temperature hyperparameter, $z_i^j = {f_p}( {z_i^j} )$, ${f_p}$ represents the projection head, and $\text{sim}(x,y) = x^{\rm{T}}y/\Vert x \Vert \Vert y \Vert$ is the $l_2$-normalized cosine similarity between $x$ and $y$. 
According to Theorem 1 in \citep{wang2020understanding}, as $N \to \infty $, Equation (\ref{eq:NCE}) can be rewritten as:
\begin{equation}\label{aassaas}
\begin{array}{l}
\scalebox{0.75}{${L_{{\rm{NCE}}}} =  - \frac{1}{\tau }\mathop {\mathbb{E}}\limits_{\left( {{z_i},{z_j}} \right) \sim {p_{pos}}} {\rm{sim}}\left( {{z_i},{z_j}} \right) + 
\mathop {\mathbb{E}}\limits_{{z^i} \sim {p_{data}}} \left[ {\log \mathop {\mathbb{E}}\limits_{{z^j} \sim {p_{data}}} \exp \left( {\frac{{{\rm{sim}}\left( {{z^i},{z^j}} \right)}}{\tau }} \right)} \right],$}
\end{array}
\end{equation}
where $p_{pos}$ represents the distribution of positive pairs and $p_{data}$ represents the data distribution. Then, based on the von Mises-Fisher (vMF) kernel density estimation (KDE) \citep{cohn2007universally, borodachov2019discrete}, the second term in Equation (\ref{aassaas}) can be considered:
\begin{equation}\label{eq:u}
\begin{array}{l}
\mathop \mathbb{E}\limits_{x \sim {p_{data}}} \left[ {\log \mathop \mathbb{E}\limits_{{x^ - } \sim {p_{data}}} \left[ {{e^{{{ - {\rm{F}}{{\left( x \right)}^{\rm{T}}}{\rm{F}}\left( {{x^ - }} \right)} \mathord{\left/
 {\vphantom {{ - {\rm{F}}{{\left( x \right)}^{\rm{T}}}{\rm{F}}\left( {{x^ - }} \right)} \tau }} \right.
 \kern-\nulldelimiterspace} \tau }}}} \right]} \right]\\
\scalebox{0.95}{$= \frac{1}{N}\sum\limits_{i = 1}^N {\log {p_{{\rm{vMF - KDE}}}}\left( {{\rm{F}}\left( {{x_i}} \right)} \right)}  + \log {Z_{{\rm{vMF}}}}\buildrel \Delta \over =  - H\left( {{\rm{F}}\left( x \right)} \right),$}
\end{array}
\end{equation}
where ${\rm F}\left( x \right) = {f_p}\left( {f\left( x \right)} \right)$, ${{p_{{\rm{vMF - KDE}}}}}$ is the KDE-based vMF kernel with $\kappa  = {\tau ^{ - 1}}$, ${Z_{{\rm{vMF}}}}$ is the vMF normalization constant for $\kappa  = {\tau ^{ - 1}}$, and $H\left(  \cdot  \right)$ is the entropy estimator.

\subsection{Generalized Learning Framework of SSCL} \label{ghjklnb}

In \textbf{Eq.(\ref{aassaas})}, the first term can be interpreted as aligning, while the second term is considered as constraining the data distribution to be uniform. From this perspective, we generalize the standard self-supervised contrastive learning (SSCL) method to a generalized learning framework (GLF).

Specifically, the proposed GLF consists of two parts: the aligning part $\mathcal{L}_{\rm{align}}$ and the constraining part $\mathcal{L}_{\rm{constrain}}$. The aligning part aims to align the two augmented samples with the same ancestor in the feature space, while the constraining part imposes a priori constraints on the training data. Thus, the objective function of GLF can be expressed as:
\begin{equation}
\mathop {\min }\limits_{f,{f_p}} {{\cal L}_{{\rm{align}}}}(X_{tr}^{aug},f,{f_p}) + {{\cal L}_{{\rm{constrain}}}}(X_{tr}^{aug},f,{f_p}).
\end{equation}

\subsection{Empirical Analysis} \label{qwwww1}

In this section, we empirically analyze four representative SSCL methods—SimCLR \citep{chen2020simple}, BYOL \citep{grill2020bootstrap}, Barlow Twins \citep{zbontar2021barlow}, and SwAV \citep{swav}—and examine the design implications for constraint components within our proposed framework.

In SimCLR, the second term of \textbf{Eq.(\ref{aassaas})} promotes a uniform distribution over augmented samples, encouraging them to span the feature space. However, this often leads to distributional overlap near the feature boundaries, reducing class separability and feature discriminability. BYOL’s objective in \textbf{Eq.(20)} constrains only the parameter update dynamics, without explicitly modeling positional relationships among augmented views. Barlow Twins (\textbf{Eq.(23)}) minimizes feature redundancy across dimensions but ignores inter-sample relationships in the embedding space. SwAV (\textbf{Eq.(26)}) introduces clustering to align augmented views but lacks mechanisms to enforce inter-class separation.
Ideally, discriminative features should exhibit clear structure: samples from the same class should cluster tightly, while those from different classes should remain well-separated. However, the above methods do not explicitly model relationships among samples with different ancestors, resulting in limited inter-class separability in the feature space.

To address this, we argue that effective constraints should explicitly encourage inter-class separation to enhance feature discriminability. We validate this claim through toy experiments on two benchmark datasets using five constraint designs:
(a) enforcing a uniform hyperspherical distribution,
(b) enforcing a uniform hypercube distribution,
(c) enforcing a Gaussian distribution with batch-wise empirical mean and covariance,
(d) enforcing a Gaussian distribution with batch-wise empirical mean and identity covariance,
(e) modeling a mixture of Gaussians by clustering each mini-batch using DBSCAN \citep{ester1996density}, computing cluster-wise means, and constructing isotropic Gaussians per cluster.
Figure \ref{fig: fig1} shows that (e) yields the best performance, suggesting that encouraging separation between samples from different categories significantly improves performance.

\subsection{Theoretical Analysis} \label{qwwww2}
In this subsection, we analyze the contrastive learning method from a theoretical perspective and provide another insight for designing the constraining part.
Given an anchor $x$, denote $x^+$ as the positive sample, $p\left( {x,{x^ + }} \right)$ as the joint distribution of $x, x^ +$, and $y \in \left\{ {1,...,K} \right\}$ as the label. Let $p(y)$ be the class probability distribution and $p_y(x)$ be the data probability distribution over class $y$. We assume positive samples are sampled from the same data distribution as an anchor and negative samples are sampled based on the probability distribution ${\mathbb{E}_{y \sim p\left( y \right)}}{p_y}\left( x \right)$.

We assume that the labels between $x$ and $x^ +$ are deterministic and consistent, e.g., $p\left( {y\left| x \right.} \right) = p\left( {y\left| {{x^ + }} \right.} \right)$. We consider the mean CE loss, which is denoted as:
\begin{equation}
\mathcal{L}_{CE}^\mu \left( f \right) = {\mathbb{E}_{p\left( {x,y} \right)}}\left[ { - \log \frac{{\exp \left( {f{{\left( x \right)}^{\rm{T}}}{\mu _y}} \right)}}{{\sum\nolimits_{i = 1}^K {\exp \left( {f{{\left( x \right)}^{\rm{T}}}{\mu _i}} \right)} }}} \right],   
\end{equation}
where ${\mu _i} = \mathbb{E}_{\left( {x\left| {y = i} \right.} \right)}\left[ {f\left( x \right)} \right]$ is the class mean vector. From \citep{arora2019theoretical}, we obtain $\mathcal{L}_{CE}^\mu \left( f \right)$ upper
bounds the standard cross entropy loss, e.g., $\mathcal{L}_{CE}^\mu \left( f \right) \ge {\min _{g_c}}{\mathcal{L}_{CE}}\left( {f,g_c} \right)$, where $g_c$ is the classifier. 
Then, let $S = \left\{ {{x_j},x_j^ + ,x_{j,1}^ - ,...,x_{j,n}^ - } \right\}_{j = 1}^M$ be the training set. We can characterize the generalization gap between contrastive learning and supervised learning risks by the following Theorem.

\begin{theorem}\label{asdadasfv}
Let ${f^*} \in \arg {\min _{f \in \mathcal{F}}}{\mathcal{L}_{\rm{NCE}}}$, where $\mathcal{F}$ represents the function space. Then with probability at least $1 - \delta $, we have:
\begin{equation}
  \mathcal{L}_{CE}^\mu \left( {{f^*}} \right) \le {\mathcal{L}_{\rm{NCE}}}\left( f \right) + \sqrt {{\rm{Var}}\left( {f\left( x \right)\left| y \right.} \right)}  + \mathcal{M}\left( n \right),
\end{equation}
where $n$ is the batch size, $\mathcal{M}\left( n \right)$ can be represented as:
\begin{equation}
\begin{array}{l}
\mathcal{M}\left( n \right) = \mathcal{O}\left( {{n^{ - {1 \mathord{\left/
 {\vphantom {1 2}} \right.
 \kern-\nulldelimiterspace} 2}}}} \right) - \log \frac{n}{K}\\
 \quad\quad\quad+ \mathcal{O}\left( {\frac{{\sqrt {1 + {1 \mathord{\left/
 {\vphantom {1 {n}}} \right.
 \kern-\nulldelimiterspace} {n }}} {\mathfrak{R} _\mathcal{F}}\left( \zeta  \right)}}{M} + \sqrt {\frac{{\log \left( {{1 \mathord{\left/
 {\vphantom {1 \delta }} \right.
 \kern-\nulldelimiterspace} \delta }} \right){{\log }^2}\left( {n } \right)}}{M}} } \right),
\end{array}
\end{equation}
where the Rademacher complexity ${\mathfrak{R}_\mathcal{F}}\left( \zeta  \right)$ is denoted as ${\mathfrak{R}_\mathcal{F}}\left( \zeta  \right) = {E_{\zeta  \sim {{\left\{ { \pm 1} \right\}}^{\left( {n + 2} \right)Md}}}}\left[ {{{\sup }_{f \in \mathcal{F}}}\left\langle {\zeta ,F} \right\rangle } \right]$, $d$ is the output dimension of the $f$ and $F \in {R^{\left( {n + 2} \right)Md}}$ can be represented as $F = 
{( {{f_t}( {{x_j}} ),{f_t}( {x_j^ + } ),{f_t}( {x_{j,1}^ - } ),...,{f_t}( {x_{j,n}^ - } )} )_{j \in [ M ],t \in [ d ]}}$.
\end{theorem}

From Theorem \ref{asdadasfv}, we observe that when ${{\rm{Var}}\left( {f\left( x \right)\left| y \right.} \right)}$ equals to 0, the $\mathcal{L}_{CE}^\mu \left( {{f^*}} \right)$ is only bounded by ${\mathcal{L}_{\rm{NCE}}}\left( f \right)$. Therefore, minimizing ${{\rm{Var}}\left( {f\left( x \right)\left| y \right.} \right)}$ is conducive to improving the performance of learned $f^*$ in downstream tasks. This further indicates
that if the constraining part under the new framework can make the points of the same category in the feature space cluster with each other, it can significantly improve the performance of SSCL methods.

\begin{figure*}[t]
    \centering
    \begin{minipage}[t]{0.6\textwidth}
        \centering
        \includegraphics[width=\textwidth]{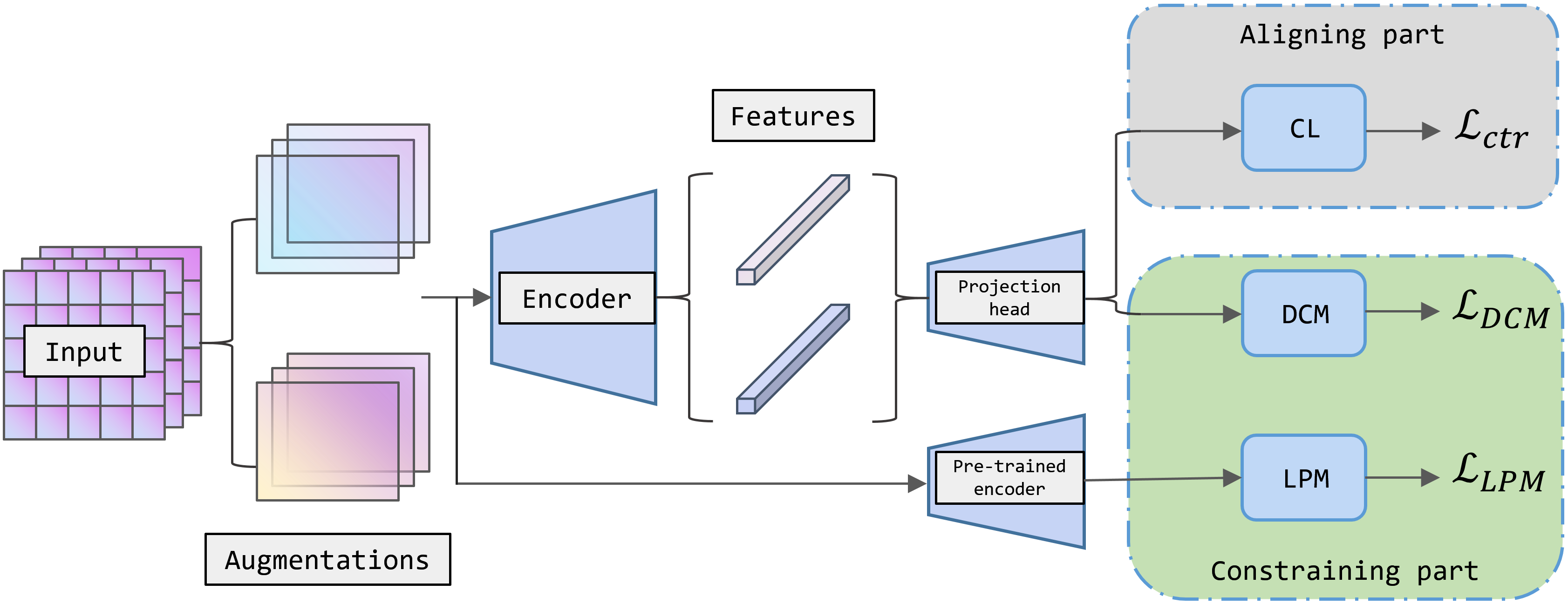}
        \caption{The overall pipeline of the proposed ADC. It consists of two parts: the aligning part and the constraining part.}
        \label{fig:fram}
    \end{minipage}%
    \hfill
    \begin{minipage}[t]{0.35\textwidth}
        \centering
        \includegraphics[width=\textwidth]{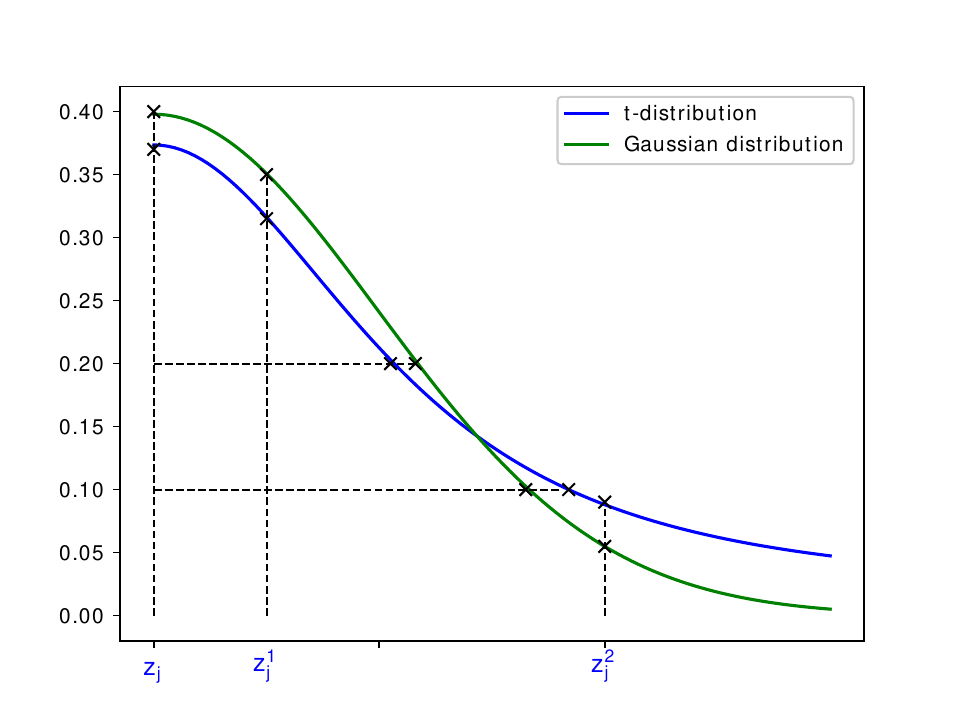}
        \caption{Visualization of the Gaussian distribution and the $t$-distribution.}
        \label{fig:gaussian}
    \end{minipage}
\end{figure*}

\section{Methodology}

Based on the above results, an ideal constraint should encourage tight clustering of same-class samples while separating those from different classes. However, without label information, class membership remains unknown during training. To address this, we propose an anchor-based strategy (\textbf{Figure~\ref{fig:fram}}), called Adaptive Distribution Calibration (ADC): each training sample is treated as an anchor, and we enforce that samples nearby in the input space stay close in the feature space, while more distant samples are pushed further apart. This process is applied iteratively across the training set. In the absence of severe outliers, this approximation promotes intra-class compactness and inter-class separability. It comprises two key components: a distribution calibration module and a local structure preservation module. Since existing SSCL methods already include an alignment objective, ADC can be used as a plug-and-play constraint module compatible with standard SSCL frameworks.

\subsection{The Proposed Method}
\subsubsection{Distribution Calibration Module} \label{sec:dcm}

The Distribution Calibration Module (DCM) dynamically induces aggregation or separation among samples in the feature space. Given an anchor, DCM attracts certain samples toward the anchor while repelling others away, thereby calibrating their relative positions. This anchor-based mechanism adjusts local feature geometry in a data-driven manner. Formally, let $p_{\text{data}}$ denote the training data distribution, and let $x_i \sim p_{\text{data}}$ be a sampled instance with feature representation $z_i = f(x_i)$. We treat $x_i$ as the anchor and apply a projection head $f_p(\cdot)$, yielding $z_i = f_p(z_i)$. The calibration distribution is modeled as a multivariate Gaussian $\mathcal{N}(z; z_i, \Sigma)$, centered at $z_i$ with covariance $\Sigma$, estimated from the training data. Meanwhile, the data distribution is approximated by a multivariate Student’s $t$-distribution $\mathcal{F}(z; z_i, \Sigma')$, with the same mean and a scaled covariance $\Sigma' = \rho \Sigma / (\rho - 2)$, where $\rho > 2$ controls the heavy-tailedness. We then define:
\begin{equation}\label{d:gs}
{\mathcal N}\left( {z;{z_i},{{\rm{\Sigma}}} } \right) = 
\frac{{\exp\{ {\textstyle{1 \over 2}}{[z - {z_i}]^{\rm{T}}}{{{\rm{\Sigma}}}^{ - 1}}[z - {z_i}]\}}}{{{\sqrt {{{( {2\pi } )}^k}\det ({\rm{\Sigma}} )} }}},
\end{equation}
where ${k}$ is the dimension of $z_i$. Meanwhile, we have:
\begin{equation}
\resizebox{0.4\textwidth}{!}{$
{\mathcal F}\left( {z;{z_i},{\rm{\Sigma}}'} \right) = \frac{{{{\rm{\Gamma} ({\textstyle{{\rho  + k} \over 2}})}}{\{ 1 + [{(z - {z_i})^{\rm{T}}}{{{\rm{\Sigma}}'}^{ - 1}}(z - {z_i})]\}^{ - {\textstyle{{\rho  + k} \over 2}}}}}}{{{{\rm{\Gamma} ({\textstyle{\rho  \over 2}}){\rho ^{{\textstyle{\rho  \over 2}}}}{\pi ^{{\textstyle{\rho  \over 2}}}}\det({\rm{\Sigma}}'){^{{\textstyle{1 \over 2}}}}}}}}
$},
\end{equation}
where $\rm{\Gamma} (  \cdot  )$ denotes the gamma function. To encourage similar samples to cluster around the anchor while pushing dissimilar samples away in the feature space, we define the objective of DCM as:
\begin{equation}\label{qw:kl}
 \mathcal{L}_{DCM} = { \sum\limits_i {{\rm{KL}}\left( {{\rm Sg}\left({\mathcal N}\left( { {{z}} ; {{z_i}} ,{\rm{\Sigma}}} \right)\right)\left| {{\mathcal F}\left( {z;z_i,{\rm{\Sigma}}'} \right)} \right.} \right)}  },
\end{equation}
where $\mathrm{KL}(\cdot)$ denotes the Kullback–Leibler divergence, and $\mathrm{Sg}(\cdot)$ is the stop-gradient operator, which prevents gradients from backpropagating through its input. Given $z_i$ and $\Sigma$, the multivariate Gaussian distribution has a higher probability density than the multivariate $t$-distribution near the mean (i.e., in the head region), but a lower density in the tails (i.e., regions far from the mean) \citep{ferguson1962representation, molenberghs1997non, van2008visualizing}. This behavior is illustrated in \textbf{Figure~\ref{fig:gaussian}}. Assuming $z^1_j$ lies close to $z_j$ and $z^2_j$ lies farther away, we derive the following ${\mathcal {N}}{(z^1_j;z_i,{\rm{\Sigma}})} > {\mathcal F}(z^1_j ;z_i,{\rm{\Sigma}}')$ and ${\mathcal N}(z^2_j;z_i,{\rm{\Sigma}}) < {\mathcal F}(z^2_j;z_i,{\rm{\Sigma}}')$.
Then, minimizing \textbf{Eq.(\ref{qw:kl})} can be thought of as constraining ${\mathcal {N}}{(z^1_j ;z_i,{\rm{\Sigma}})} = {\mathcal F}(z^1_j ;z_i,{\rm{\Sigma}}')$ and ${\mathcal N}(z^2_j;z_i,{\rm{\Sigma}}) = {\mathcal F}(z^2_j;z_i,{\rm{\Sigma}}')$.
If the density value of each sample in ${\mathcal {N}}{(\cdot ;z_i,{\rm{\Sigma}})}$ is fixed, the way for minimizing \textbf{Eq.(\ref{qw:kl})} is to restrict $z^1_j$ and $z_i$ to be closer together while $z^2_j$ and $z_i$ are farther apart in the feature space, thus leading to larger ${\mathcal F}(z^1_j ;z_i,{\rm{\Sigma}}')$ and smaller ${\mathcal F}(z^2_j;z_i,{\rm{\Sigma}}')$.

Empirically, given a mini-batch of training data $X = \left\{ {{x_1},...,{x_n}} \right\}$ with batch size $n$, we input them into functions $f$ and $f_p$ to obtain $Z = \left\{ {{z_1},...,{z_n}} \right\}$. Using $z_i$ as the anchor, we present the discretized forms of ${\mathcal N}(z_j;{z_i},{\rm{\Sigma}})$ and ${\mathcal F}(z_j;z_i,{\rm{\Sigma}}')$ respectively. For ${z_j} \in Z$, the calibration distribution values $p_{cal}^i$ and data distribution values $p_{dat}^i$ is:
\begin{equation}\label{eq:dsds}
\begin{array}{l}
{p_{cal}^i\left( z_j \right)} = \frac{{\mathcal{N}\left( { {{z_j}} ;{{z_i}} ,{\rm{\Sigma}}} \right)}}{{\sum\limits_{j = 1}^n {\mathcal{N}\left( { {{z_j}} ; {{z_i}} ,{\rm{\Sigma}}} \right)} }},{p_{dat}^i\left( {z_j} \right)} = \frac{{\mathcal{F}\left( {{z_j};{z_i},{\rm{\Sigma}}'} \right)}}{{\sum\limits_{j = 1}^n {\mathcal{F}\left( {{z_j};{z_i},{\rm{\Sigma}}'} \right)} }}.
\end{array}
\end{equation}
Therefore, \textbf{Eq.(\ref{qw:kl})} can be rewritten as:
\begin{equation}\label{eq:1212sd}
{\mathcal{L}_{DCM}} = \sum\limits_{i = 1}^n {\sum\limits_{j = 1}^{n-1}  {{\rm Sg}\left(p_{cal}^i\left( {{z_j}} \right) \right) \log \frac{{\rm Sg}\left({p_{cal}^i\left( {{z_j}} \right)}\right)}{{p_{dat}^i\left( {{z_j}} \right)}}} }.
\end{equation}
By minimizing \textbf{Eq.(\ref{eq:1212sd})}, samples with high probability density (i.e., close to the anchor in feature space) are pulled closer to the anchor, while those with low density (i.e., farther from the anchor) are pushed away. As nearby samples are more likely to belong to the same class, iterating over the entire training set and treating each sample as an anchor enables approximate distribution calibration, promoting intra-class compactness and inter-class separability.

\begin{figure}[t]
    \centering
    \subfigure[$H(\boldsymbol{\alpha}) =0.32 $]{\includegraphics[width=0.15\textwidth]{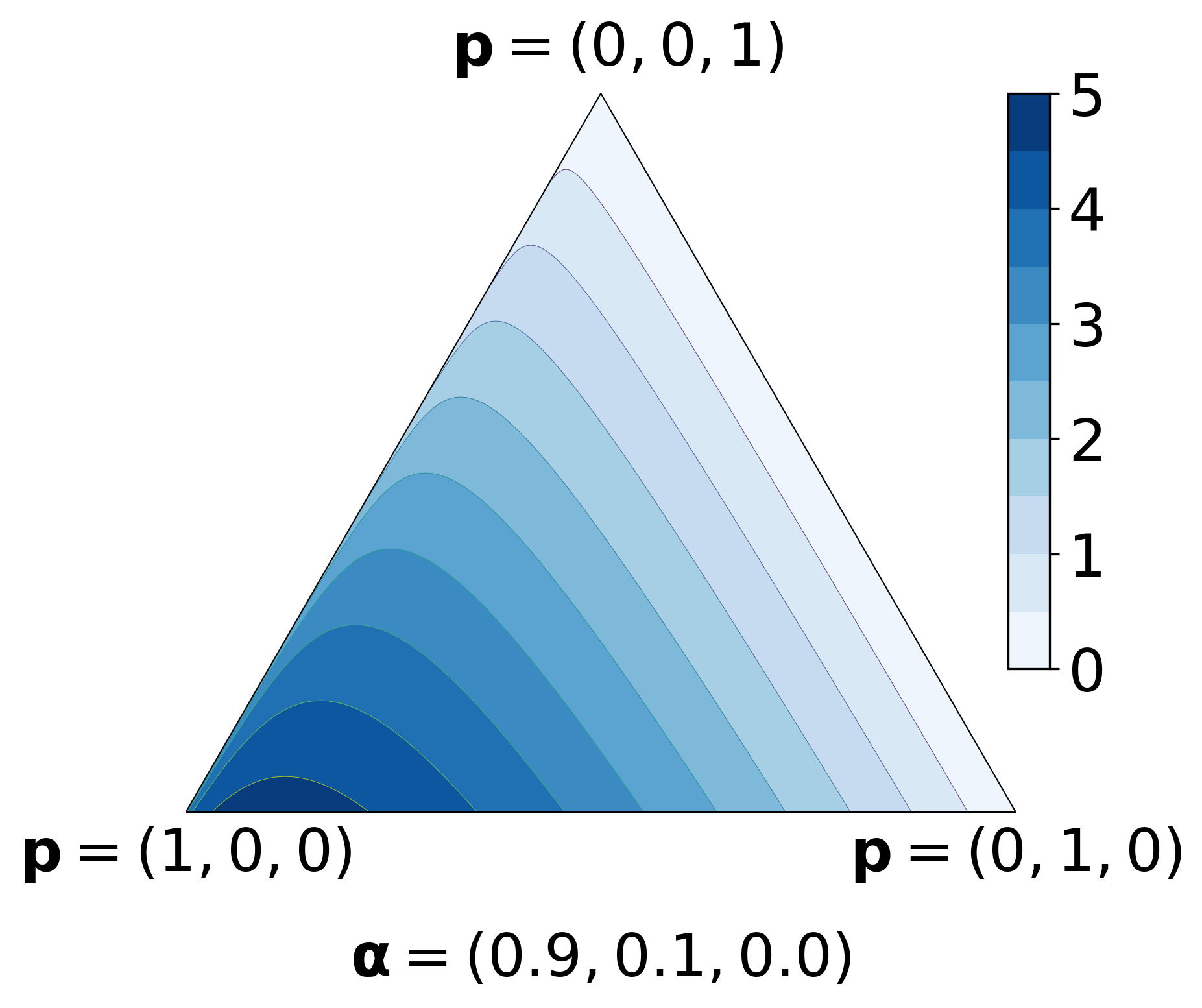}\label{fig: d2}}
    \subfigure[$H(\boldsymbol{\alpha}) =0.41 $]{\includegraphics[width=0.15\textwidth]{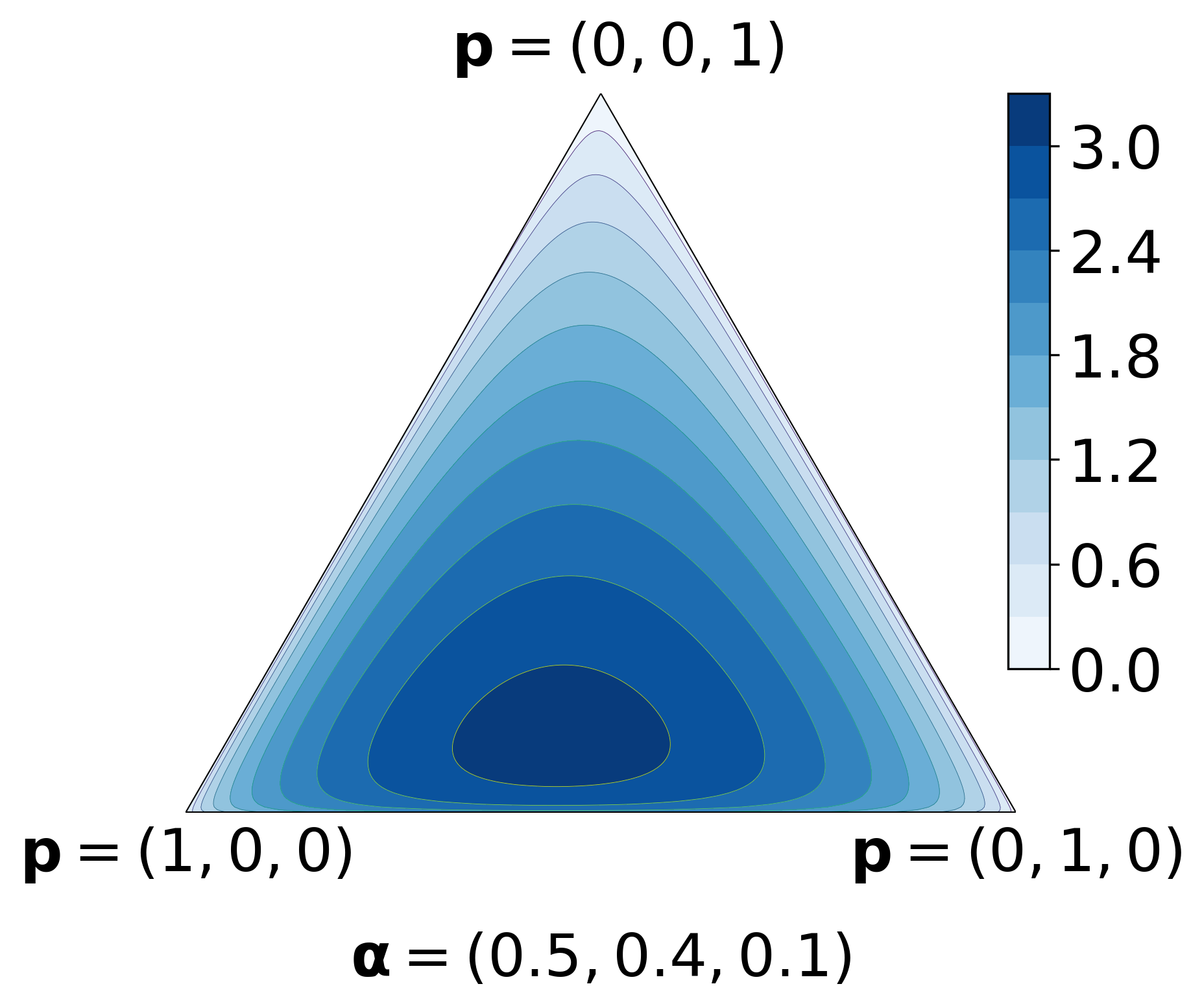}\label{fig: d3}}
    \subfigure[$H(\boldsymbol{\alpha}) =1.08 $]{\includegraphics[width=0.15\textwidth]{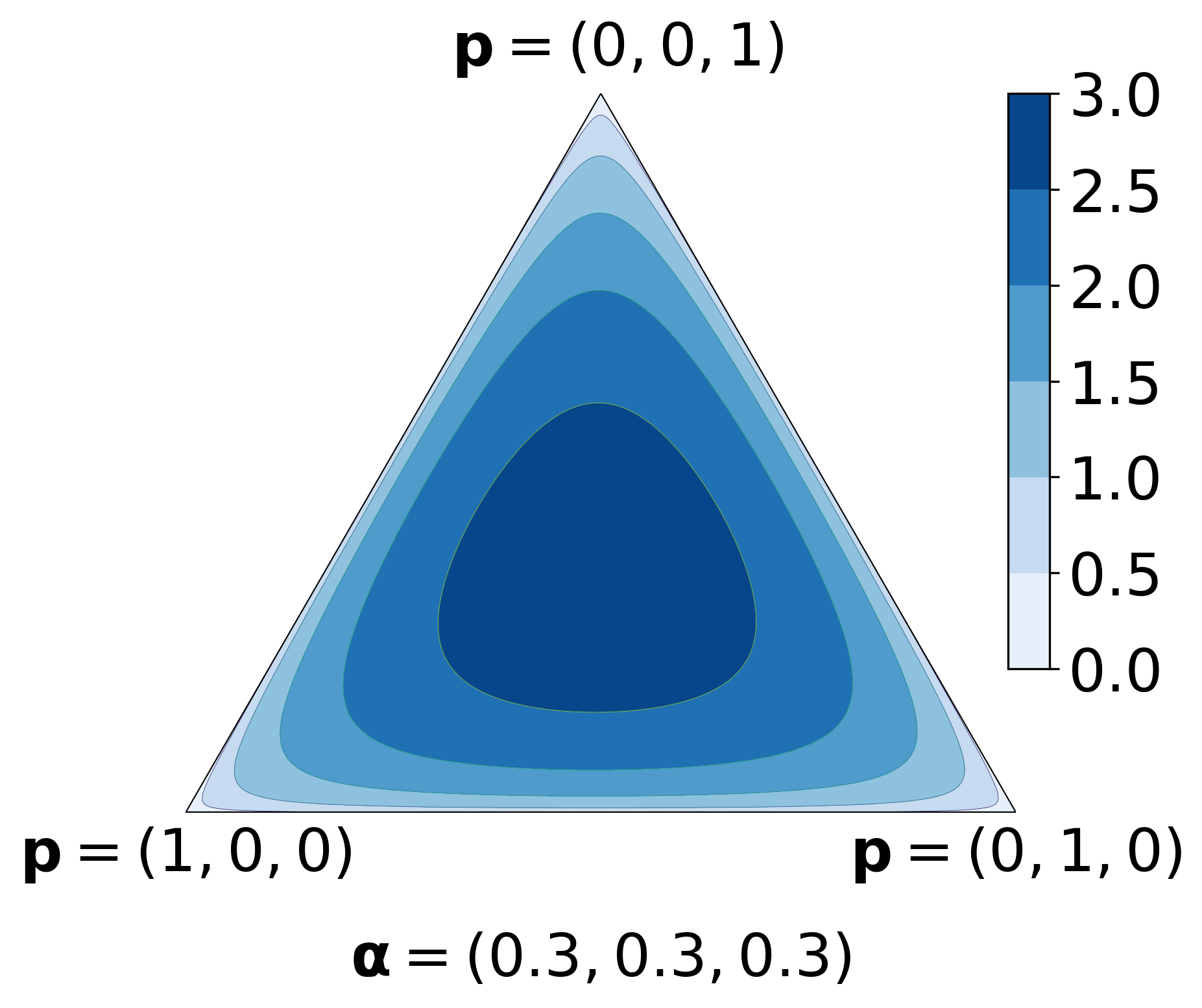}\label{fig: d1}}
    \caption{The visualizations of three kinds of DD.}
    \label{fig: dsdsfig1}
\end{figure}

\subsubsection{Local Preserving Module}
\label{sec:lpm}
DCM operates in the feature space by measuring distances between augmented samples, but its effectiveness may be compromised during early training when the feature extractor is under-optimized. In this phase, samples that are close or distant in the input space may be misrepresented in the feature space. Moreover, if an outlier is chosen as the anchor, DCM may erroneously attract surrounding samples, leading to distorted representations. To address these issues, we introduce a Local Preservation Module (LPM) that enforces local consistency by preserving relative input-space distances in the feature space. LPM also quantifies the likelihood of an anchor being an outlier using information entropy and down-weights its influence through an adaptive weighting scheme. Before presenting DCM in detail, we first introduce the Dirichlet Distribution (DD) \citep{ng2011dirichlet}.
\begin{definition}\textbf{(Dirichlet Distribution)}
	\label{def:dd}
	The Dirichlet distribution of order $K \ge 2$ with parameters ${\alpha _1},...,{\alpha _K} \ge 0$ has a probability density function with respect to Lebesgue measure on the Euclidean space ${\mathbb{R}^{K - 1}}$, which can be given by
	\begin{equation}\label{qw123}
	Dir\left( {\boldsymbol{\beta} \left| \boldsymbol{\alpha}  \right.} \right) = \left\{ \begin{array}{l}
\frac{1}{{\rm{B}\left( \boldsymbol{\alpha}  \right)}}\prod _{i = 1}^K\beta _i^{{\alpha _i} - 1},{\rm{ for }} \, \boldsymbol{\beta}  \in {\boldsymbol{S}_K}\\
0,{\rm{ otherwise}},
\end{array} \right.
	\end{equation}
	where $\boldsymbol{\alpha} = \left[ {{\alpha _1},...,{\alpha _K}} \right]$, $\boldsymbol{\beta} = \left[ {{\beta _1},...,{\beta _K}} \right]$, and ${\boldsymbol{S}_K}$ belong to the standard $k-1$ simplex, or in other words ${S_K} = \left\{ {\boldsymbol{\beta} \left| {\sum\limits_{i = 1}^K {{\beta _i}}  = 1{\rm{ \,and\, 0}} \le {\beta _1},...,{\beta _K} \le 1} \right.} \right\}$.
	The normalizing constant is the multivariate beta function, which can be expressed in terms of the gamma function ${\rm{B}}\left( {\boldsymbol{\alpha}} \right) = {{\mathop \prod \limits_{i = 1}^K \rm{\Gamma} \left( {{\alpha _{\rm{i}}}} \right)} \mathord{\left/
 {\vphantom {{\mathop \prod \limits_{i = 1}^K \rm{\Gamma} \left( {{\alpha _{\rm{i}}}} \right)} {\rm{\Gamma} \left( {\sum\limits_{{\rm{i = 1}}}^{\rm{K}} {{\alpha _{\rm{i}}}} } \right)}}} \right.
 \kern-\nulldelimiterspace} {{\rm{\Gamma}} \left( {\sum\limits_{{{i = 1}}}^{K} {{\alpha _{{i}}}} } \right)}}$,

where $\rm{\Gamma}(\cdot)$ is the gamma function.	
	
\end{definition}
According to \textbf{Eq.(\ref{qw123})}, the expectation of DD is:
\begin{equation}
    {\mathbb{E}_{Dir\left( {\boldsymbol{\beta} \left| \boldsymbol{\alpha}  \right.} \right)}}\left( \boldsymbol{\beta}  \right) = {\boldsymbol{\alpha} \mathord{\left/
 {\vphantom {a {\sum\nolimits_{i = 1}^K {{a_i}} }}} \right.
 \kern-\nulldelimiterspace} {\sum\nolimits_{i = 1}^K {{a_i}} }}.
\end{equation}
We have $\mathbb{E}_{\mathrm{Dir}(\boldsymbol{\beta} \mid \boldsymbol{\alpha})}(\boldsymbol{\beta}) = \boldsymbol{\alpha}$ when $\sum_{i=1}^K a_i = 1$. Based on Definition 1, the probability density of $\mathrm{Dir}(\boldsymbol{\beta} \mid \boldsymbol{\alpha})$ is high when $\boldsymbol{\beta}$ is close to $\boldsymbol{\alpha}$, and low otherwise. As illustrated in Figure~~\ref{fig: dsdsfig1}, the Dirichlet distribution exhibits an aggregation behavior: when $\sum_{i=1}^K a_i = 1$, a smaller entropy $H(\boldsymbol{\alpha})$ leads to a sharper concentration of the probability density near $\boldsymbol{\alpha}$, while a larger entropy flattens the density surface, spreading probability mass more evenly across the domain.

The Local Preservation Module (LPM) is derived from this property of the Dirichlet distribution. Given a mini-batch of training data $X = \{x_1, \dots, x_n\}$ with corresponding feature representations $Z = \{z_1, \dots, z_n\}$ computed by functions $f$ and $f_p$, we use a CLIP-pretrained feature extractor $f_{\text{pre}}$ \citep{RadfordKHRGASAM21} to obtain prior representations $Z^{\text{pre}} = \{z^{\text{pre}}_1, \dots, z^{\text{pre}}_n\}$. Treating $x_i$ as the anchor, its prior representation is $z^{\text{pre}}_i$, and the corresponding multivariate $t$-distribution is denoted $\mathcal{F}(z^{\text{pre}}; z^{\text{pre}}_i, \Sigma')$. The normalized, discretized form of this distribution is given by:
\begin{equation}\label{eq:prot}
{p_{pre}^i\left( z^{pre}_j \right)} = \frac{{\mathcal{F}\left( {z^{pre}_j;z^{pre}_i,{{\rm{\Sigma}}'}} \right)}}{{\sum\limits_{j = 1}^n {\mathcal{F}\left( {z^{pre}_j;z^{pre}_i,{{\rm{\Sigma}}'}} \right)} }}, \forall z_j^{pre} \in {Z^{pre}}.
\end{equation}
We define $\mathbb{P}_{\text{data}}^i = \{ p_{\text{data}}^i(z_j) \}_{j \neq i}$ and $\mathbb{P}_{\text{pre}}^i = \{ p_{\text{pre}}^i(z_j) \}_{j \neq i}$, excluding the anchor index $i$. Since $\mathbb{P}_{\text{pre}}^i$ is computed using a pretrained model known for producing semantically meaningful features, distances in $Z^{\text{pre}}$ better reflect the true relative positions. Thus, samples closer to the anchor have higher probabilities in $\mathbb{P}_{\text{pre}}^i$, and more distant ones have lower values. To preserve this, LPM encourages $\mathbb{P}_{\text{data}}^i$ to lie in the high-density region of $\mathrm{Dir}(\mathbb{P}_{\text{data}}^i \mid \mathbb{P}_{\text{pre}}^i)$, yielding:
\begin{equation}\label{dfddfd}
    {\mathcal{L}_{LPM}} = \sum\limits_{i=1}^{n} {{Dir\left( {{\mathbb{P}^{data}_i}\left| {{\mathbb{P}^{pre}_i}} \right.} \right)}}.
\end{equation}

Maximizing \textbf{Eq.~(\ref{dfddfd})} encourages samples that are close in the input space to remain close in the feature space, while pushing apart those that are distant, thereby mitigating DCM's miscalibration. We do not enforce $\mathbb{P}^{\text{pre}}_i = \mathbb{P}^{\text{cal}}_i$ explicitly, as anchors are randomly selected from $X$, and the induced multivariate $t$-distribution accurately reflects local structure only when the anchor aligns with the true class center. Ideally, samples from the same class would share a common anchor. Optimizing \textbf{Eq.~(\ref{dfddfd})} implicitly promotes consistent local topology across anchors or alignment with class centers. Moreover, when an anchor is an outlier, its distances to other samples tend to be uniform, causing $\mathbb{P}^{\text{pre}}_i$ to approximate a uniform distribution. To reduce the impact of such cases, LPM incorporates a weighted calibration strategy into \textbf{Eq.~(\ref{eq:1212sd})}, leveraging properties of the Dirichlet distribution:
\begin{equation}\label{eq:121dasdad2sd}
\begin{array}{l}
{\mathcal{L}_{DCM}} = \frac{1}{{H\left( {{\mathbb{P}^{pre}_i}} \right)}}\sum\limits_{i = 1}^n {\sum\limits_{j = 1}^{n-1}  {{\rm Sg}\left(p_{cal}^i\left( {{z_j}} \right) \right) \log \frac{{\rm Sg}\left({p_{cal}^i\left( {{z_j}} \right)}\right)}{{p_{dat}^i\left( {{z_j}} \right)}}} }.
\end{array}
\end{equation}
As observed from the properties of the Dirichlet distribution, when the anchor lies within the distribution, the induced $\mathbb{P}^{\text{pre}}_i$ contains a mix of high and low probabilities, resulting in low entropy $H(\mathbb{P}^{\text{pre}}_i)$. In contrast, when the anchor is an outlier, the probabilities are nearly uniform, yielding high entropy. Therefore, minimizing \textbf{Eq.~(\ref{eq:121dasdad2sd})} effectively reduces the influence of outlier anchors.

\subsection{Overall Objective}

Finally, given the feature extractor $f$, the pre-trained $f_{pre}$, the projection head $f_p$, the objective of ADC is:
\begin{equation}\label{eq:sssb}
\begin{array}{l}
\scalebox{0.92}{$\mathcal{L} \left( {f,f_{p}} \right)= 
{\mathcal{L}_{ctr}}\left( {f, f_{p}} \right) + \nu{\mathcal{L}_{DCM}}\left( {f, f_{p}} \right) - \upsilon{\mathcal{L}_{LPM}\left( {f, f_{p}} \right)},$}
\end{array}
\end{equation}
where $\nu, \upsilon > 0$ are temperature hyperparameters, and $\mathcal{L}_{\text{ctr}}(f, f_p)$ denotes the loss used in SSCL methods. Unlike negative-sample-based methods (NSB-m) such as Hard \citep{robinson2020contrastive} and Debiased \citep{chuang2020debiased}, which mitigate false negatives by pulling together augmented views of the same instance, ADC fundamentally differs in its objective. NSB-m methods often ignore relationships between different instances of the same class, resulting in high within-class variance. In contrast, ADC explicitly promotes intra-class compactness by aligning samples from different instances but the same class, while ensuring inter-class separation across distinct ancestors, thereby enhancing both alignment and discrimination.

\begin{table}[t]
\centering
\small
\setlength{\tabcolsep}{1mm}
\begin{tabular}{lcccccc}
\toprule
\multirow{2.5}{*}{Method} & \multicolumn{2}{c}{CIFAR-10} & \multicolumn{2}{c}{CIFAR-100} & \multicolumn{2}{c}{STL-10} \\
\cmidrule(lr){2-3} \cmidrule(lr){4-5} \cmidrule(lr){6-7}
& \textbf{linear} & \textbf{5-nn} & \textbf{linear} & \textbf{5-nn} & \textbf{linear} & \textbf{5-nn} \\
\midrule
MoCo & 91.69 & 88.66 & 67.02 & 56.29 & 90.64 & 88.01 \\
SimSiam & 91.71 & 88.65 & 67.22 & 56.36 & 91.01 & 88.16 \\
SimCLR & 91.80 & 88.42 & 66.83 & 56.56 & 90.51 & 85.68 \\
BYOL & 91.73 & 89.45 & 66.60 & 56.82 & 91.99 & 88.64 \\
Barlow Twins & 91.43 & 89.68 & 66.13 & 56.70 & 90.38 & 87.13 \\
W-MSE & 91.99 & 89.87 & 67.64 & 56.45 & 91.75 & 88.59 \\
ReSSL & 90.20 & 88.26 & 66.79 & 53.72 & 88.25 & 86.33 \\
MEC & 90.55 & 87.80 & 67.36 & 57.25 & 91.33 & 89.03 \\
VICRegL & 90.99 & 88.75 & 68.03 & 57.34 & 92.12 & 90.01 \\
\midrule
SimSiam+DCM & 92.21 & 89.25 & 67.89 & 57.56 & 91.61 & 88.69 \\
SimCLR+DCM & 92.34 & 89.05 & 67.52 & 57.10 & 91.12 & 86.51 \\
BYOL+DCM & 92.30 & 90.02 & 67.20 & 57.38 & 92.51 & 89.51 \\
Barlow Twins+DCM & 91.97 & 90.22 & 66.83 & 57.29 & 91.01 & 87.77 \\
W-MSE+DCM & 92.58 & 90.39 & 68.22 & 57.01 & 92.27 & 89.11 \\
VICRegL+DCM & 92.00 & 90.15 & 68.38 & 57.32 & 91.89 & 89.06 \\
\midrule
SimSiam+LPM & 92.22 & 89.19 & 67.91 & 57.48 & 91.64 & 88.72 \\
SimCLR+LPM & 92.35 & 89.11 & 67.44 & 57.18 & 91.21 & 86.47 \\
BYOL+LPM & 92.31 & 90.01 & 67.16 & 57.31 & 92.53 & 89.49 \\
Barlow Twins+LPM & 91.94 & 90.21 & 66.81 & 57.30 & 91.09 & 87.74 \\
W-MSE+LPM & 92.54 & 90.42 & 68.27 & 57.11 & 92.21 & 89.02 \\
VICRegL+LPM & 91.95 & 90.19 & 68.42 & 57.28 & 91.92 & 89.01 \\
\midrule
SimSiam+ADC & 93.41 & 90.62 & 69.23 & 58.89 & 92.86 & 89.80 \\
SimCLR+ADC & 93.46 & 89.55 & 68.93 & 58.51 & 92.35 & 88.02 \\
BYOL+ADC & 93.70 & 89.35 & 68.43 & 58.72 & 94.01 & 90.91 \\
Barlow Twins+ADC & 93.19 & 90.32 & 68.24 & 58.62 & 92.20 & 89.14 \\
W-MSE+ADC & 93.81 & 91.52 & 69.57 & 58.29 & 93.68 & 90.43 \\
VICRegL+ADC & 92.84 & 90.52 & 69.04 & 59.02 & 92.18 & 89.07 \\
\bottomrule
\end{tabular}
\caption{The classification accuracies of a linear classifier (linear) and a 5-nearest neighbors classifier (5-nn) with a ResNet-18. Full results are shown in \textbf{Table 6}.}
\label{tab:1_part}
\end{table}

\begin{table*}[t]  
	\centering	
	\begin{tabular}{lcccccccc}
   
		\toprule

\multirow{2}{*}{Method} & \multicolumn{4}{c}{CIFAR-100} & \multicolumn{4}{c}{ImageNet} \\

\cmidrule(lr){2-5} 
\cmidrule(lr){6-9}

& $f_{pre}^{\rm I}$ & $f_{pre}^{\rm SimCLR}$ & $f_{pre}^{\rm CLIP}$ & $f_{pre}^{\rm EVA-02}$ 
& $f_{pre}^{\rm I}$ & $f_{pre}^{\rm SimCLR}$ & $f_{pre}^{\rm CLIP}$ & $f_{pre}^{\rm EVA-02}$ \\
   
\midrule

SimCLR + ADC & 68.94 & 68.93 & 68.93 & 68.92 & 73.21 & 73.22 & 77.12 & 77.12 \\
BYOL + ADC & 68.43 & 68.45 & 68.43 & 68.44 & 77.12 & 77.13 & 77.12 & 77.12 \\
Barlow Twins + ADC & 68.24 & 68.23 & 68.24 & 68.25 & 76.05 & 76.03 & 76.03 & 76.04 \\

		\bottomrule
	\end{tabular}
	\caption{The comparison results using CIFAR-100 and ImageNet across different baselines and pre-trained models.}
	\label{tab:pret}
\end{table*}

\begin{table}
	\centering
    \small
	\begin{tabular}{lccccc}
		\toprule
		\multirow{2.5}{*}{Method} 
		& \multicolumn{2}{c}{VOC 07} 
		& \multicolumn{2}{c}{COCO} \\
	    \cmidrule(lr){2-3} \cmidrule(lr){4-5} 
	    & \( \mathbf{AP_{50}} \) & \( \mathbf{AP} \)
	    & \( \mathbf{AP^{mask}_{50}} \) & \( \mathbf{AP^{mask}} \) \\
	    \midrule
	    Supervised & 74.4 & 42.4 & 54.7 & 33.3 \\
	    \midrule
	    SimCLR & 75.9 & 46.8 & 54.6 & 33.3 \\
	    MoCo & 77.1 & 46.8 & 55.8 & 34.4 \\
	    SimSiam & 77.3 & 48.5 & 56.0 & 34.4 \\
	    MEC & 77.4 & 48.3 & 56.3 & 34.7 \\
	    RELIC v2 & 76.9 & 48.0 & 56.0 & 34.6 \\
	    CorInfoMax & 76.8 & 47.6 & 56.2 & 34.8 \\
	    VICRegL & 75.9 & 47.4 & 56.5 & 35.1 \\
	    \midrule
	    SimCLR + ADC & 77.5 & 49.0 & 56.3 & 35.5 \\
	    MoCo + ADC & 79.0 & \textbf{50.3} & 57.6 & 36.0 \\
	    SimSiam + ADC & \textbf{79.0} & 50.3 & 57.6 & 35.9 \\
	    MEC + ADC & 78.9 & 50.2 & \textbf{57.9} & \textbf{36.1} \\
	    VICRegL + ADC & 78.0 & 49.3 & 56.6 & 36.3 \\
		\bottomrule
	\end{tabular}
    \caption{Transfer learning results on VOC 07 detection and COCO instance segmentation using C4-backbone.}
	\label{tab:voc_coco_only}
\end{table}

\begin{figure}[t]
  \centering
  \begin{minipage}{0.23\textwidth}
    \centering
    \includegraphics[width=\textwidth]{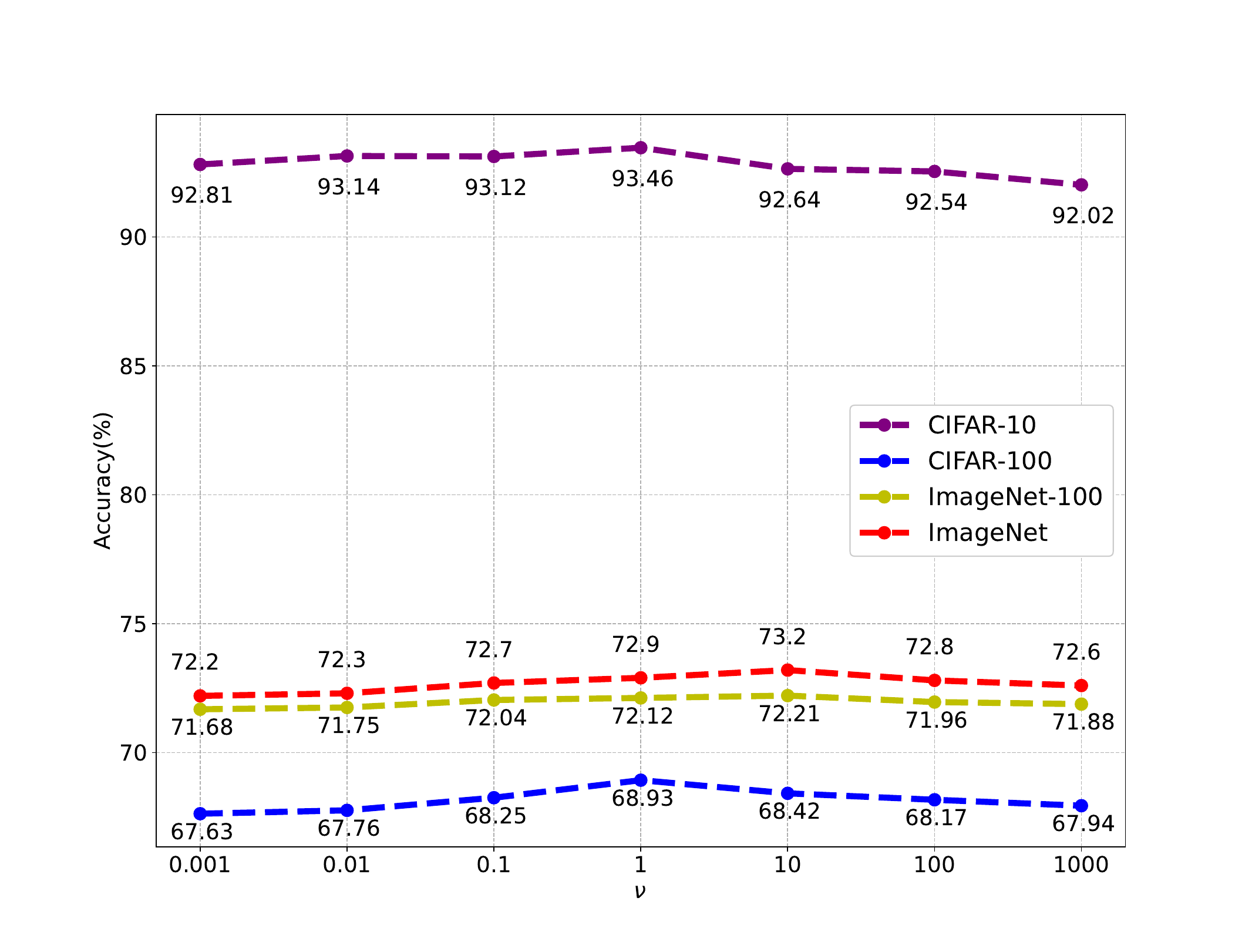}
    \caption{Evaluation on $\nu$.}
    \label{fig:param_1}
  \end{minipage}
  \hfill
  \begin{minipage}{0.23\textwidth}
    \centering
    \includegraphics[width=\textwidth]{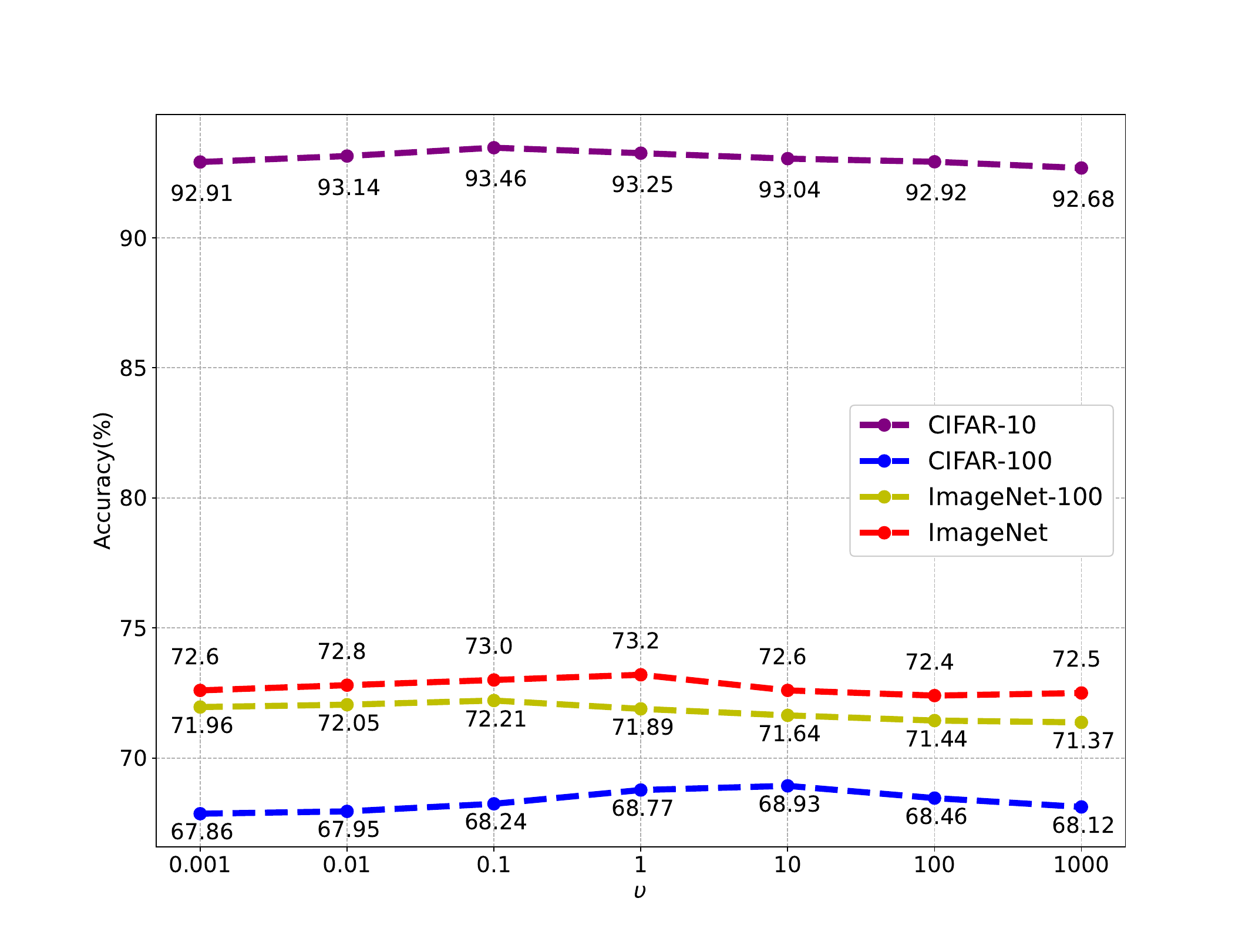}
    \caption{Evaluation on $\upsilon$.}
    \label{fig:param_2}
  \end{minipage}
\end{figure}

\section{Experiments}
To evaluate ADC, we conduct extensive experiments on both the benchmark and the real-world datasets, including tasks like classification, object detection, and segmentation (See \textbf{Appendices} for details and full results). All experiments are the average of five runs on NVIDIA V100 GPUs. 

\subsection{Comprison and Evaluation}

\subsubsection{Benchmark Dataset} \label{dadadadad}

We select on eight benchmark datasets, including CIFAR-10 \citep{cifar10}, CIFAR-100 \citep{cifar10}, STL-10 \citep{stl10}, Tiny ImageNet \citep{leTinyImagenetVisual2015}, ImageNet-100 \citep{tianContrastiveMultiviewCoding2020}, ImageNet \citep{krizhevsky2012imagenet}, PASCAL VOC \citep{everingham2010pascal}, and COCO \citep{lin2014microsoft}. For real-world data, we deploy a radar- and camera-based workstation near the coast of Sanya, operating continuously to monitor offshore regions and capture large volumes of ship imagery.  
The collected raw data is then processed into two structured datasets. 
\textbf{Full results and details are provided in Appendix due to space limitation.}

\subsubsection{Unsupervised learning}
Following \cite{chen2020simple}, we freeze the pretrained encoder and train a supervised linear classifier on top. The classifier is optimized using SGD (momentum 0.9, weight decay $5\times10^{-6}$) for 500 epochs with a learning rate decaying from $10^{-2}$ to $10^{-6}$. 
\textbf{Table~\ref{tab:1_part}} shows that our ADC-based methods consistently outperform existing baselines. \textbf{Table 6} presents top-1 and top-5 accuracy on ImageNet-100 using ResNet-50, also demonstraing advantages. More results are shown in \textbf{Appendix}.

\subsubsection{Semi-supervised learning}
We fine-tune models on 1\% and 10\% subsets of ImageNet labels. Spatial augmentations are applied during training, and standard resizing, cropping, and normalization are used during testing.
\textbf{Table 8} shows that ADC consistently improves top-1 and top-5 accuracy across both settings, improving by more than 3.5\%.

\subsubsection{Transfer learning}
Following \citep{chen2020simple,liu2022self}, we evaluate on object detection and instance segmentation using PASCAL VOC and COCO datasets. We adopt Faster R-CNN with a C4 backbone for VOC and Mask R-CNN for COCO, using Detectron2 settings with learning rate search. \textbf{Table 3} shows that SSL methods generally match or exceed supervised baselines, and achieve the best performance across all tasks with ADC.

\subsubsection{Evaluation on real-world application} 
We conduct experiments on the Real Ship Classification Dataset (RSCD) and the Real Ship Detection Dataset (RSDD), which includes 10 classes, e.g., cargo ships, fishing boats, sailboats, yachts, etc. The collected images cover diverse scenarios, including challenging conditions, e.g., strong reflections, occlusions, and small targets. \textbf{Figure 8}, \textbf{Table 10}, and \textbf{Table 11} demonstrate that ADC consistently achieves best results.

\subsection{Ablation study}
We conduct multiple ablation studies and analyses, e.g., the two modules of ADC, feature extractor, hyperparameters, ADC under supervised setting, robustness to outliers, and feature visualization (see Appendix for full results).

\subsubsection{Influence of the proposed two modules.}
By setting $\upsilon = 0$ and $\nu = 0$, we remove the influence of LPM and DCM. The results in \textbf{Tables \ref{tab:1_part}, 7, and 4} show that while DCM and LPM individually outperform baselines, the full ADC model achieves significantly better performance.

\subsubsection{Influence of the pre-trained feature extractor}\label{asdgjlgy}
We select multiple pre-trained settings:
(i) an identity matrix $f_{\text{pre}}^{\rm I}$ for direct similarity computation,
(ii) SimCLR pretrained on ImageNet ($f_{\text{pre}}^{\rm SimCLR}$),
(iii) the CLIP visual encoder ($f_{\text{pre}}^{\rm CLIP}$) \citep{RadfordKHRGASAM21}, and
(iv) the EVA-02 visual encoder ($f_{\text{pre}}^{\rm EVA-02}$) \citep{corrabs}.
\textbf{Table 2} shows that ADC is robust to the choice of pre-trained models.

\subsubsection{Parameter Sensitivity} 
We investigate the influence of hyperparameters in ADC, i.e., $\nu$ and $\upsilon$, with the range of $\left\{ {{{10}^{ - 3}},{{10}^{ - 2}},{{10}^{ - 1}},1,10,{{10}^2},{{10}^3}} \right\}$ on multiple benchmark datasets. \textbf{Figures 5 and 6} show that the optimal setting for different datasets and demonstrate its stability in practice.

\section{Conclusion}
In this paper, we propose a Generalized Learning Framework (GLF) for SSCL, which has shown impressive performance on various downstream tasks. GLF consists of an aligning component that encourages positive pairs to have similar representations, and a constraining component that imposes additional regularization on the feature space. We analyze representative SSCL methods under GLF based on quantitative measures of their aligning and constraining effects. Based on our analysis, we identify two desirable properties of the constraining component: it should maximize the intra-class compactness and inter-class separability of the features. To achieve these properties, we introduce a plug-and-play method, which adapts the feature distribution to match the input distribution. Extensive theoretical and empirical results demonstrate that ADC achieves stable improvement on both benchmarks and real-world data.

\bibliography{aaai2026}

\begin{thebibliography}{45}
\providecommand{\natexlab}[1]{#1}

\bibitem[{Arora et~al.(2019)Arora, Khandeparkar, Khodak, Plevrakis, and Saunshi}]{arora2019theoretical}
Arora, S.; Khandeparkar, H.; Khodak, M.; Plevrakis, O.; and Saunshi, N. 2019.
\newblock A theoretical analysis of contrastive unsupervised representation learning.
\newblock \emph{arXiv preprint arXiv:1902.09229}.

\bibitem[{Borodachov, Hardin, and Saff(2019)}]{borodachov2019discrete}
Borodachov, S.~V.; Hardin, D.~P.; and Saff, E.~B. 2019.
\newblock \emph{Discrete energy on rectifiable sets}.
\newblock Springer.

\bibitem[{Caron et~al.(2020)Caron, Misra, Mairal, Goyal, Bojanowski, and Joulin}]{swav}
Caron, M.; Misra, I.; Mairal, J.; Goyal, P.; Bojanowski, P.; and Joulin, A. 2020.
\newblock Unsupervised Learning of Visual Features by Contrasting Cluster Assignments.
\newblock \emph{CoRR}, abs/2006.09882.

\bibitem[{Chen et~al.(2021)Chen, Niu, Gong, Li, Yang, and Sugiyama}]{chen2021large}
Chen, S.; Niu, G.; Gong, C.; Li, J.; Yang, J.; and Sugiyama, M. 2021.
\newblock Large-margin contrastive learning with distance polarization regularizer.
\newblock In \emph{International Conference on Machine Learning}, 1673--1683. PMLR.

\bibitem[{Chen et~al.(2020{\natexlab{a}})Chen, Kornblith, Norouzi, and Hinton}]{chen2020simple}
Chen, T.; Kornblith, S.; Norouzi, M.; and Hinton, G. 2020{\natexlab{a}}.
\newblock A simple framework for contrastive learning of visual representations.
\newblock In \emph{International conference on machine learning}, 1597--1607. PMLR.

\bibitem[{Chen, Luo, and Li(2021)}]{chen2021intriguing}
Chen, T.; Luo, C.; and Li, L. 2021.
\newblock Intriguing properties of contrastive losses.
\newblock \emph{Advances in Neural Information Processing Systems}, 34: 11834--11845.

\bibitem[{Chen et~al.(2020{\natexlab{b}})Chen, Fan, Girshick, and He}]{chen2020improved}
Chen, X.; Fan, H.; Girshick, R.; and He, K. 2020{\natexlab{b}}.
\newblock Improved baselines with momentum contrastive learning.
\newblock \emph{arXiv preprint arXiv:2003.04297}.

\bibitem[{Chen and He(2021)}]{chen2021exploring}
Chen, X.; and He, K. 2021.
\newblock Exploring simple siamese representation learning.
\newblock In \emph{Proceedings of the IEEE/CVF Conference on Computer Vision and Pattern Recognition}, 15750--15758.

\bibitem[{Chen, Xie, and He(2021)}]{chen2021empirical}
Chen, X.; Xie, S.; and He, K. 2021.
\newblock An empirical study of training self-supervised vision transformers.
\newblock In \emph{Proceedings of the IEEE/CVF International Conference on Computer Vision}, 9640--9649.

\bibitem[{Chuang et~al.(2020)Chuang, Robinson, Lin, Torralba, and Jegelka}]{chuang2020debiased}
Chuang, C.-Y.; Robinson, J.; Lin, Y.-C.; Torralba, A.; and Jegelka, S. 2020.
\newblock Debiased contrastive learning.
\newblock \emph{Advances in neural information processing systems}, 33: 8765--8775.

\bibitem[{Coates, Ng, and Lee(2011)}]{stl10}
Coates, A.; Ng, A.; and Lee, H. 2011.
\newblock An analysis of single-layer networks in unsupervised feature learning.
\newblock In \emph{Proceedings of the fourteenth international conference on artificial intelligence and statistics}, 215--223. JMLR Workshop and Conference Proceedings.

\bibitem[{Cohn and Kumar(2007)}]{cohn2007universally}
Cohn, H.; and Kumar, A. 2007.
\newblock Universally optimal distribution of points on spheres.
\newblock \emph{Journal of the American Mathematical Society}, 20(1): 99--148.

\bibitem[{Ermolov et~al.(2021)Ermolov, Siarohin, Sangineto, and Sebe}]{ermolov2021whitening}
Ermolov, A.; Siarohin, A.; Sangineto, E.; and Sebe, N. 2021.
\newblock Whitening for self-supervised representation learning.
\newblock In \emph{International Conference on Machine Learning}, 3015--3024. PMLR.

\bibitem[{Everingham et~al.(2010)Everingham, Van~Gool, Williams, Winn, and Zisserman}]{everingham2010pascal}
Everingham, M.; Van~Gool, L.; Williams, C.~K.; Winn, J.; and Zisserman, A. 2010.
\newblock The pascal visual object classes (voc) challenge.
\newblock \emph{International journal of computer vision}, 88(2): 303--338.

\bibitem[{Fang et~al.(2023)Fang, Sun, Wang, Huang, Wang, and Cao}]{corrabs}
Fang, Y.; Sun, Q.; Wang, X.; Huang, T.; Wang, X.; and Cao, Y. 2023.
\newblock {EVA-02:} {A} Visual Representation for Neon Genesis.
\newblock \emph{CoRR}, abs/2303.11331.

\bibitem[{Ferguson(1962)}]{ferguson1962representation}
Ferguson, T.~S. 1962.
\newblock A representation of the symmetric bivariate Cauchy distribution.
\newblock \emph{The Annals of Mathematical Statistics}, 33(4): 1256--1266.

\bibitem[{Grill et~al.(2020{\natexlab{a}})Grill, Strub, Altch{\'e}, Tallec, Richemond, Buchatskaya, Doersch, Avila~Pires, Guo, Gheshlaghi~Azar et~al.}]{grill2020bootstrap}
Grill, J.-B.; Strub, F.; Altch{\'e}, F.; Tallec, C.; Richemond, P.; Buchatskaya, E.; Doersch, C.; Avila~Pires, B.; Guo, Z.; Gheshlaghi~Azar, M.; et~al. 2020{\natexlab{a}}.
\newblock Bootstrap your own latent-a new approach to self-supervised learning.
\newblock \emph{Advances in neural information processing systems}, 33: 21271--21284.

\bibitem[{Grill et~al.(2020{\natexlab{b}})Grill, Strub, Altch{\'e}, Tallec, Richemond, Buchatskaya, Doersch, Pires, Guo, Azar et~al.}]{byol}
Grill, J.-B.; Strub, F.; Altch{\'e}, F.; Tallec, C.; Richemond, P.~H.; Buchatskaya, E.; Doersch, C.; Pires, B.~A.; Guo, Z.~D.; Azar, M.~G.; et~al. 2020{\natexlab{b}}.
\newblock Bootstrap your own latent: A new approach to self-supervised learning.
\newblock \emph{arXiv preprint arXiv:2006.07733}.

\bibitem[{He et~al.(2020)He, Fan, Wu, Xie, and Girshick}]{moco}
He, K.; Fan, H.; Wu, Y.; Xie, S.; and Girshick, R. 2020.
\newblock Momentum contrast for unsupervised visual representation learning.
\newblock In \emph{Proceedings of the IEEE/CVF conference on computer vision and pattern recognition}, 9729--9738.

\bibitem[{Hjelm et~al.(2018)Hjelm, Fedorov, Lavoie-Marchildon, Grewal, Bachman, Trischler, and Bengio}]{hjelm2018learning}
Hjelm, R.~D.; Fedorov, A.; Lavoie-Marchildon, S.; Grewal, K.; Bachman, P.; Trischler, A.; and Bengio, Y. 2018.
\newblock Learning deep representations by mutual information estimation and maximization.
\newblock \emph{arXiv preprint arXiv:1808.06670}.

\bibitem[{Jaiswal et~al.(2020)Jaiswal, Babu, Zadeh, Banerjee, and Makedon}]{jaiswal2020survey}
Jaiswal, A.; Babu, A.~R.; Zadeh, M.~Z.; Banerjee, D.; and Makedon, F. 2020.
\newblock A survey on contrastive self-supervised learning.
\newblock \emph{Technologies}, 9(1): 2.

\bibitem[{Krizhevsky, Hinton et~al.(2009)}]{cifar10}
Krizhevsky, A.; Hinton, G.; et~al. 2009.
\newblock Learning multiple layers of features from tiny images.

\bibitem[{Krizhevsky, Sutskever, and Hinton(2012)}]{krizhevsky2012imagenet}
Krizhevsky, A.; Sutskever, I.; and Hinton, G.~E. 2012.
\newblock Imagenet classification with deep convolutional neural networks.
\newblock \emph{Advances in neural information processing systems}, 25: 1097--1105.

\bibitem[{Le and Yang(2015)}]{leTinyImagenetVisual2015}
Le, Y.; and Yang, X. 2015.
\newblock Tiny Imagenet Visual Recognition Challenge.
\newblock \emph{CS 231N}, 7(7): 3.

\bibitem[{Li et~al.(2022)Li, Qiang, Zheng, Su, and Xiong}]{li2022metaug}
Li, J.; Qiang, W.; Zheng, C.; Su, B.; and Xiong, H. 2022.
\newblock MetAug: Contrastive Learning via Meta Feature Augmentation.
\newblock \emph{International Conference on Machine Learning}.

\bibitem[{Li et~al.(2020)Li, Zhou, Xiong, and Hoi}]{li2020prototypical}
Li, J.; Zhou, P.; Xiong, C.; and Hoi, S.~C. 2020.
\newblock Prototypical contrastive learning of unsupervised representations.
\newblock \emph{arXiv preprint arXiv:2005.04966}.

\bibitem[{Lin et~al.(2014)Lin, Maire, Belongie, Hays, Perona, Ramanan, Doll{\'a}r, and Zitnick}]{lin2014microsoft}
Lin, T.-Y.; Maire, M.; Belongie, S.; Hays, J.; Perona, P.; Ramanan, D.; Doll{\'a}r, P.; and Zitnick, C.~L. 2014.
\newblock Microsoft coco: Common objects in context.
\newblock In \emph{European conference on computer vision}, 740--755. Springer.

\bibitem[{Liu et~al.(2022)Liu, Wang, Li, and Wang}]{liu2022self}
Liu, X.; Wang, Z.; Li, Y.; and Wang, S. 2022.
\newblock Self-Supervised Learning via Maximum Entropy Coding.
\newblock \emph{arXiv preprint arXiv:2210.11464}.

\bibitem[{Molenberghs and Lesaffre(1997)}]{molenberghs1997non}
Molenberghs, G.; and Lesaffre, E. 1997.
\newblock Non-linear integral equations to approximate bivariate densities with given marginals and dependence function.
\newblock \emph{Statistica Sinica}, 713--738.

\bibitem[{Ng, Tian, and Tang(2011)}]{ng2011dirichlet}
Ng, K.~W.; Tian, G.-L.; and Tang, M.-L. 2011.
\newblock Dirichlet and related distributions: Theory, methods and applications.

\bibitem[{Oord, Li, and Vinyals(2018)}]{oord2018representation}
Oord, A. v.~d.; Li, Y.; and Vinyals, O. 2018.
\newblock Representation learning with contrastive predictive coding.
\newblock \emph{arXiv preprint arXiv:1807.03748}.

\bibitem[{Radford et~al.(2021{\natexlab{a}})Radford, Kim, Hallacy, Ramesh, Goh, Agarwal, Sastry, Askell, Mishkin, Clark, Krueger, and Sutskever}]{RadfordKHRGASAM21}
Radford, A.; Kim, J.~W.; Hallacy, C.; Ramesh, A.; Goh, G.; Agarwal, S.; Sastry, G.; Askell, A.; Mishkin, P.; Clark, J.; Krueger, G.; and Sutskever, I. 2021{\natexlab{a}}.
\newblock Learning Transferable Visual Models From Natural Language Supervision.
\newblock In Meila, M.; and Zhang, T., eds., \emph{Proceedings of the 38th International Conference on Machine Learning, {ICML} 2021, 18-24 July 2021, Virtual Event}, volume 139 of \emph{Proceedings of Machine Learning Research}, 8748--8763. {PMLR}.

\bibitem[{Radford et~al.(2021{\natexlab{b}})Radford, Kim, Hallacy, Ramesh, Goh, Agarwal, Sastry, Askell, Mishkin, Clark et~al.}]{radford2021learning}
Radford, A.; Kim, J.~W.; Hallacy, C.; Ramesh, A.; Goh, G.; Agarwal, S.; Sastry, G.; Askell, A.; Mishkin, P.; Clark, J.; et~al. 2021{\natexlab{b}}.
\newblock Learning transferable visual models from natural language supervision.
\newblock In \emph{International Conference on Machine Learning}, 8748--8763. PMLR.

\bibitem[{Ram et~al.(2010)Ram, Jalal, Jalal, and Kumar}]{ester1996density}
Ram, A.; Jalal, S.; Jalal, A.~S.; and Kumar, M. 2010.
\newblock A density based algorithm for discovering density varied clusters in large spatial databases.
\newblock \emph{International Journal of Computer Applications}, 3(6): 1--4.

\bibitem[{Robinson et~al.(2020)Robinson, Chuang, Sra, and Jegelka}]{robinson2020contrastive}
Robinson, J.; Chuang, C.-Y.; Sra, S.; and Jegelka, S. 2020.
\newblock Contrastive learning with hard negative samples.
\newblock \emph{arXiv preprint arXiv:2010.04592}.

\bibitem[{Si et~al.(2022)Si, Qiang, Li, Xu, and Sun}]{SI2022727}
Si, L.; Qiang, W.; Li, J.; Xu, F.; and Sun, F. 2022.
\newblock Multi-view representation learning from local consistency and global alignment.
\newblock \emph{Neurocomputing}, 501: 727--740.

\bibitem[{Tian, Krishnan, and Isola(2020{\natexlab{a}})}]{tian2020contrastive}
Tian, Y.; Krishnan, D.; and Isola, P. 2020{\natexlab{a}}.
\newblock Contrastive multiview coding.
\newblock In \emph{European conference on computer vision}, 776--794. Springer.

\bibitem[{Tian, Krishnan, and Isola(2020{\natexlab{b}})}]{tianContrastiveMultiviewCoding2020}
Tian, Y.; Krishnan, D.; and Isola, P. 2020{\natexlab{b}}.
\newblock Contrastive {{Multiview Coding}}.
\newblock arXiv:1906.05849.

\bibitem[{Tomasev et~al.(2022)Tomasev, Bica, McWilliams, Buesing, Pascanu, Blundell, and Mitrovic}]{tomasev2022pushing}
Tomasev, N.; Bica, I.; McWilliams, B.; Buesing, L.; Pascanu, R.; Blundell, C.; and Mitrovic, J. 2022.
\newblock Pushing the limits of self-supervised ResNets: Can we outperform supervised learning without labels on ImageNet?
\newblock \emph{arXiv preprint arXiv:2201.05119}.

\bibitem[{Van~der Maaten and Hinton(2008)}]{van2008visualizing}
Van~der Maaten, L.; and Hinton, G. 2008.
\newblock Visualizing data using t-SNE.
\newblock \emph{Journal of machine learning research}, 9(11).

\bibitem[{Wang et~al.(2023)Wang, Mou, Ma, Huang, and Gao}]{wang2023amsa}
Wang, J.; Mou, L.; Ma, L.; Huang, T.; and Gao, W. 2023.
\newblock AMSA: Adaptive multimodal learning for sentiment analysis.
\newblock \emph{ACM Transactions on Multimedia Computing, Communications and Applications}, 19(3s): 1--21.

\bibitem[{Wang et~al.(2024)Wang, Mou, Zheng, and Gao}]{wang2024image}
Wang, J.; Mou, L.; Zheng, C.; and Gao, W. 2024.
\newblock Image-based freeform handwriting authentication with energy-oriented self-supervised learning.
\newblock \emph{IEEE Transactions on Multimedia}.

\bibitem[{Wang and Isola(2020)}]{wang2020understanding}
Wang, T.; and Isola, P. 2020.
\newblock Understanding contrastive representation learning through alignment and uniformity on the hypersphere.
\newblock In \emph{International Conference on Machine Learning}, 9929--9939. PMLR.

\bibitem[{Zbontar et~al.(2021)Zbontar, Jing, Misra, LeCun, and Deny}]{zbontar2021barlow}
Zbontar, J.; Jing, L.; Misra, I.; LeCun, Y.; and Deny, S. 2021.
\newblock Barlow twins: Self-supervised learning via redundancy reduction.
\newblock \emph{arXiv preprint arXiv:2103.03230}.

\bibitem[{Zheng et~al.(2021)Zheng, You, Wang, Qian, Zhang, Wang, and Xu}]{zheng2021ressl}
Zheng, M.; You, S.; Wang, F.; Qian, C.; Zhang, C.; Wang, X.; and Xu, C. 2021.
\newblock Ressl: Relational self-supervised learning with weak augmentation.
\newblock \emph{Advances in Neural Information Processing Systems}, 34: 2543--2555.

\end{thebibliography}

\end{document}